\documentclass[runningheads]{llncs}
\usepackage[T1]{fontenc}
\usepackage{graphicx}
\usepackage{booktabs}
\usepackage{hyperref}
\usepackage{color}

\begin{document}
\title{A Synthetic Benchmark to Explore Limitations of Localized Drift Detections}
%
%



\author{
Flavio Giobergia\inst{1} \and 
Eliana Pastor\inst{1} \and 
Luca de Alfaro\inst{2} \and
Elena Baralis \inst{1}
}
\authorrunning{F. Giobergia et al.}
%
\institute{
Politecnico di Torino, Italy \\
\email{\{firstname.lastname\}@polito.it}\and
University of California, Santa Cruz, USA\\
\email{luca@ucsc.edu}
}
\maketitle              
\begin{abstract}
Concept drift is a common phenomenon in data streams where the statistical properties of the target variable change over time. Traditionally, drift is assumed to occur globally, affecting the entire dataset uniformly. However, this assumption does not always hold true in real-world scenarios where only specific subpopulations within the data may experience drift. This paper explores the concept of localized drift and evaluates the performance of several drift detection techniques in identifying such localized changes. We introduce a synthetic dataset based on the Agrawal generator, where drift is induced in a randomly chosen subgroup. Our experiments demonstrate that commonly adopted drift detection methods may fail to detect drift when it is confined to a small subpopulation. We propose and test various drift detection approaches to quantify their effectiveness in this localized drift scenario. We make the source code for the generation of the synthetic benchmark available at \url{https://github.com/fgiobergia/subgroup-agrawal-drift}.

\keywords{Drift detection  \and Synthetic data \and Localized drift.}
\end{abstract}
\section{Introduction}
In the realm of data stream mining, the detection of concept drift is of fundamental importance to maintain the accuracy and reliability of predictive models.
Concept drift refers to the change in the statistical properties of the target variable that the model is trying to predict.
Traditionally, drift detection techniques make the (often implicit) assumption that the drift occurs globally, i.e., the change is uniformly distributed across the entire dataset.
This assumption, however, may not always hold in real-world situations where drift can occur in a localized manner, affecting only certain subpopulations within the data (e.g., only young women employed in the IT sector).

\begin{figure}
\centering
\includegraphics[width=\linewidth]{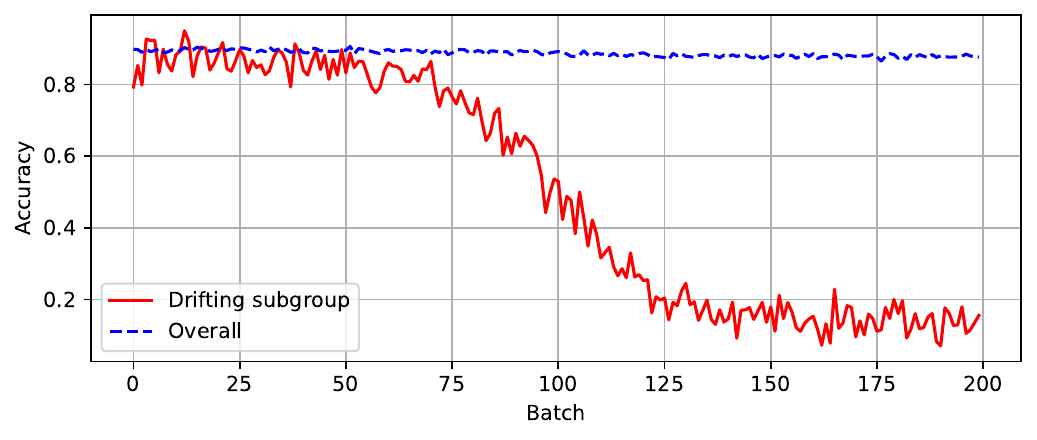}
\label{fig:example}
\caption{Accuracy computed on the overall dataset and on the drifting subgroup (2\% of the dataset), throughout a drifting event.}

\end{figure}

Localized drift poses a significant challenge for traditional drift detection methods. 
These methods are designed to identify global changes and may overlook drifts that are confined to a small subset of the data.
As a result, models may fail to adapt to these local changes, leading to degraded performance and inaccurate predictions.
For instance, a subgroup covering 2\% of the population may start behaving in a significantly different way than previously known. 
It is desirable that this change in behavior be detected by drift detectors. 
However, the drift goes unnoticed when we observe the overall performance of the model (i.e., the performance on the entire population).
Figure~\ref{fig:example} shows how the accuracy varies under subgroup drift for the entire population and for the specific subgroup.
While the overall performance degrades by approximately 2\% and may go unnoticed, the accuracy within the subgroup drops to 0. 

To investigate the limitations of existing drift detection methods in the context of localized drift, we introduce a synthetic dataset inspired by the Agrawal generator~\cite{agrawal1993database}.
In this dataset, drift is intentionally induced in a randomly chosen subgroup of a specific size, while the rest of the data remains stable.
This setup allows us to simulate a scenario where only a specific subpopulation is subject to drift, thereby providing a controlled environment to evaluate the effectiveness of various drift detection techniques.

The primary contributions of this paper are as follows:
\begin{itemize}
    \item We highlight the importance of recognizing localized drift in data streams and its implications for drift detection methodologies.
    \item We introduce a synthetic dataset based on the Agrawal generator with induced localized drift, providing a benchmark for evaluating drift detection methods.
    \item We conduct a comprehensive evaluation of several drift detection techniques, quantifying their performance in detecting localized drift.
\end{itemize}

The rest of the paper is organized as follows. Section~\ref{sec:dataset} describes the synthetic dataset and experimental setup. Section~\ref{sec:results} presents the results of our experiments and discusses the findings. Finally, Section~\ref{sec:conclusions} concludes the paper and suggests directions for future research.

\section{Proposed dataset}
\label{sec:dataset}
We introduce a novel dataset based on the synthetic one proposed in~\cite{agrawal1993database}. In particular, we propose (i) identifying a randomly selected subgroup of the population, defined as a slice of the dataset's attributes and of a user-specified size, and (ii) only injecting this target subgroup with noise to simulate a situation where the drift occurs locally, instead of globally. The code is available at \url{https://github.com/fgiobergia/subgroup-agrawal-drift}.

\subsection{Subgroup Agrawal Drift Dataset}
\label{ssec:agrawal}
To explore the concept of localized drift, we define a synthetic dataset based on the Agrawal generator~\cite{agrawal1993database}. The Agrawal generator is commonly used for simulating data streams and generates samples $x$ in a domain $\mathcal{D}$ with six numerical attributes and three categorical attributes, producing binary classification tasks. The attributes are as follows:
\begin{itemize}
    \item \texttt{salary}, uniformly distributed from \$20,000 to \$150,000
    \item \texttt{commission}, 0 if \texttt{salary} has a value below \$75,000, otherwise it is uniformly distributed from \$10,000 to \$75,000
    \item \texttt{age}, uniformly distributed from 20 to 80
    \item \texttt{elevel} (education level), uniformly chosen from 0 to 4
    \item \texttt{car} (car maker), uniformly chosen from 1 to 20
    \item \texttt{zipcode} (zip code of the town), uniformly chosen from 0 to 8
    \item \texttt{hvalue} (house value), uniformly distributed from $\$50,000 \,\cdot\,$ \texttt{zipcode} to\\ $\$100,000 \,\cdot\,$ \texttt{zipcode}. Different zip codes, as such, are associated with different average house prices
    \item \texttt{hyears} (years the house has been owned), 1 to 30 uniformly distributed  
    \item \texttt{loan} (total loan amount requested), uniformly distributed from \$0 to \$500,000
\end{itemize}

Each synthetic record is associated with a binary outcome (i.e., whether the loan is approved or not).
Ten different functions, $f_0(x), f_1(x), \cdots, f_9(x)$ have been proposed in the original work to map a given record $x$ to the binary ground truth value, $f_i : \mathcal{D} \rightarrow \{0, 1\}$.
This completely defines any record $\{ x, f_k(x) \}$ used either during training or in deployment.
A perturbation can also be included so as not to make the classification task trivial.
This perturbation (ranging from 0\% to 100\%) affects the attributes of $x$ \textit{after} the class has been assigned, thus adding an element of fuzzyness in the sample/class relationship. 

A common technique to introduce concept drift \cite{montiel2021river} consists of adopting a classification function $f_i$ for the original concept and a different one $f_j$ ($i \ne j$) for the drift concept.
At step $t$, the function is defined as a random variable $F$:

\begin{equation}
    F = Z \cdot f_i(x) + (1 - Z) \cdot f_j(x),
\end{equation}
where $Z \sim Bernoulli(p_t)$. In this way, $F(x) = f_i(x)$ with probability $p_t$, $f_j(x)$ otherwise.
The drift occurs gradually over time by defining $p_t$ through a sigmoid function, $p_t = (1 + e^{-4(t-k)/w})^{-1}$. $k$ represents the center of the sigmoid function, and $w$ is its width. 

This drift, however, is applied uniformly to all samples. Instead, we aim to create a drift that is localized in nature, i.e., that only affects one subpopulation of the dataset. We consider a selector function $s : \mathcal{D} \rightarrow \{0, 1\}$, having value 1 for samples belonging to the target subgroup, 0 otherwise. We outline the definition of $s(\cdot)$ in more detail in Subsection \ref{ssec:subgroup-def}. We define $F$ as follows:
\begin{equation}
    F = s(x) \cdot [Z \cdot f_i(x) + (1 - Z) \cdot f_j(x)] + [1 - s(x)] f_i(x).
\end{equation}
In other words, the gradual drift is applied only to samples belonging to the target subgroup defined by $s(\cdot)$. All other samples will retain the original concept.  We note that this definition can be extended to multiple subpopulations, which can be subject to different drifts. 

\subsection{Subgroup definition}
\label{ssec:subgroup-def}
To simulate a localized drift, we need to define a target subgroup within the dataset. We produce meaningful subgroups by identifying slices of the domain, e.g., \{ age $\in$ [25, 30], salary $\in$ [\$75,000, \$100,000] \}.  We fully automate the synthetic dataset generation phase by introducing a random subgroup definition policy. This policy produces, for a desired subgroup size (i.e., subgroup support), a slice of the population that approximately encompasses it. 

\begin{figure}[hbt!]
     \centering
     \includegraphics[width=.75\linewidth]{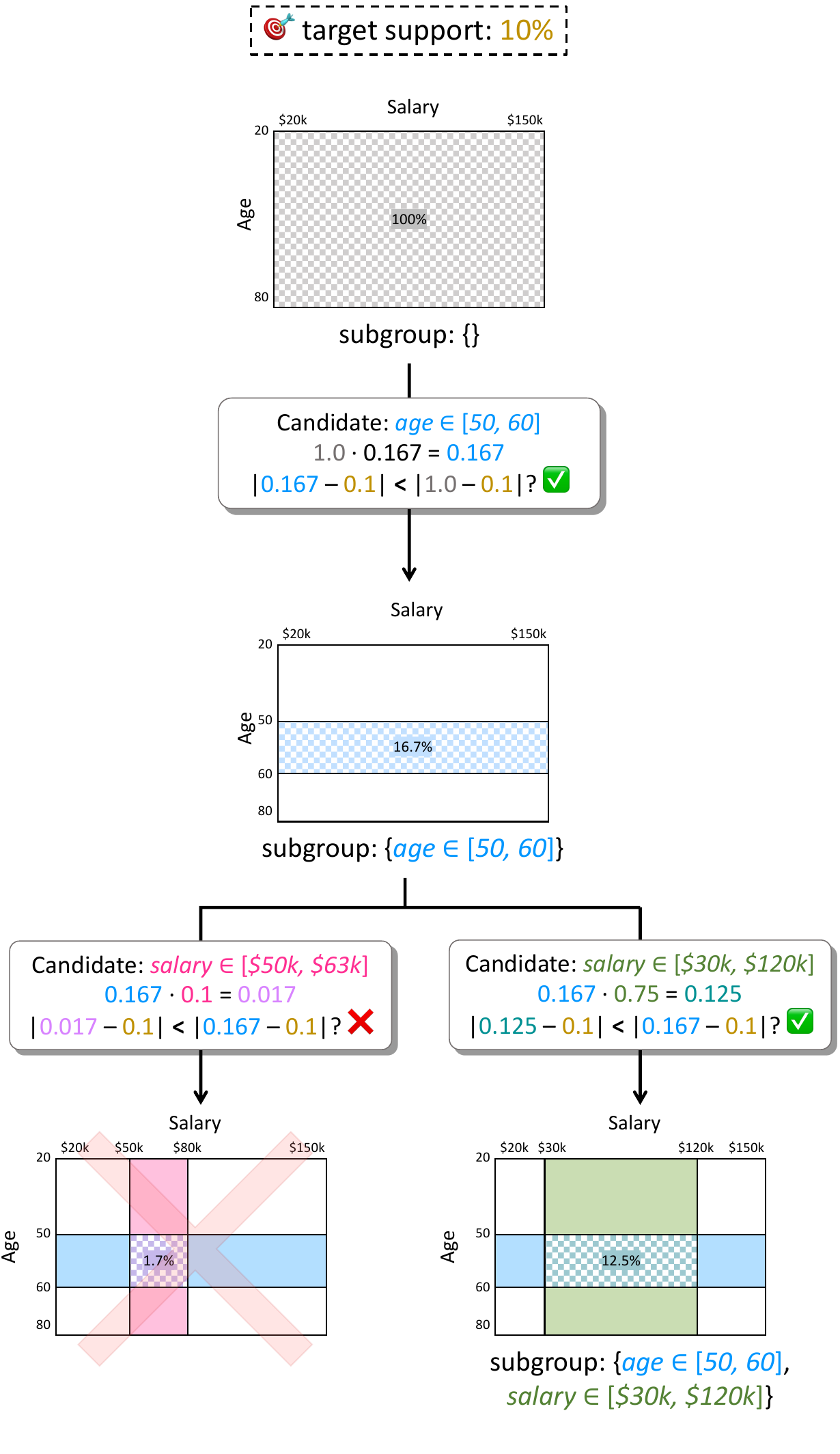}
     \caption{Example of the greedy process adopted to randomly generate subgroups on 2 attributes. From top to bottom, the target subgroup is built by iteratively adding randomly generated slices of the attributes, if their inclusion produces a better approximation of the desired support (in the example, 10\%). The checkered area at each step represents the size (support) of the current subgroup. }
     \label{fig:example-support}
 \end{figure}

We adopt a greedy policy to identify a subset of slices that, combined, well approximate the target subgroup size. We do this by identifying random ranges of values (e.g., $[c, d]$) for randomly chosen attributes (e.g., $\texttt{attr} \sim U(a, b)$, $a \le c < d \le b$). The uniform distribution makes it trivial to compute the probability of belonging to the random range of values, as $P(\texttt{attr} \in [c, d]) = P(\texttt{attr}) = \frac{d-c}{b-a}$. Additionally, the attributes are independent from one another\footnote{As discussed in Subsection \ref{ssec:agrawal}, all but two attributes (\texttt{commission} and \texttt{hvalue}) are independently sampled from uniform distributions with known ranges. We only consider independent attributes for the definition of the target subgroup for simplicity.}. As such, their combined probability can be computed as the product of the separate probabilities, $P(\texttt{attr.1})\cdot P(\texttt{attr.2})\cdot \ldots \cdot P(\texttt{attr.n})$. We either include or discard a candidate slice based on whether it gets the current probability closer to the target one. Figure \ref{fig:example-support} provides an example where a subgroup of approximately the target size (10\%) is iteratively defined by identifying a first slice on \texttt{age}, followed by a second one on \texttt{salary}. 
 Because of the greedy nature of the algorithm, slices that do not provide an immediate improvement in terms of support are discarded.
 The algorithm terminates when either the subgroup size is within a tolerance threshold of the target one, or a maximum number of iterations is reached. 

\subsection{Examples of generated subgroups}
In this subsection, we provide some instances of generated subgroups for various target subgroup sizes.

\begin{figure}
    \centering
    \includegraphics[width=\linewidth]{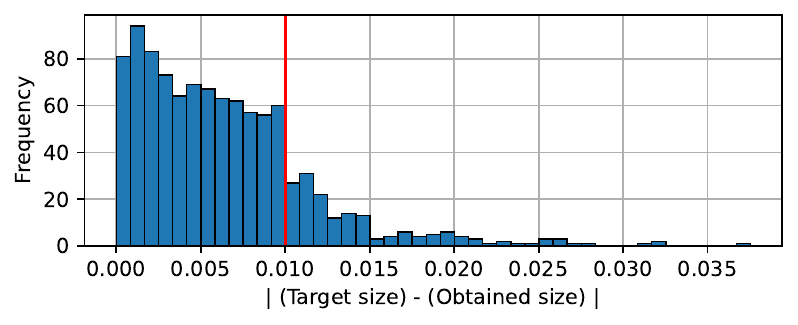}
    \caption{Distribution of the absolute difference between target and corresponding obtained subgroup sizes, for 1,000 generated subgroups and various target sizes, tolerance of 0.01 (marked in red), maximum number of iterations set to 1,000.}
    \label{fig:delta}
\end{figure}

To guarantee variety in the generated subgroups, the proposed generator produces random subgroups that approximate the desired target size. As detailed above, the algorithm refines the generated subgroup until either the desired tolerance is reached or a maximum number of iterations has been executed. 
Figure~\ref{fig:delta} shows the distribution of gaps between target and obtained subgroup sizes, for a tolerance of 0.01 on the target size. In 83\% of cases, the desired tolerance is reached, whereas in the remaining 17\% of cases the maximum number of iterations (1,000) is reached. 

We report four examples of generated subgroups in Table \ref{tab:subgroup-examples}. For each, we report the target (desired) size (5\%, 10\%, 25\% and 50\% of the population, respectively), the one computed according to the greedy policy adopted for subgroup generation, and the actual (empirical) size, as observed over a generated sample of 10,000 points. Both computed and actual sizes are close to the target one. If needed, the gap between computed and target subgroup sizes can be lowered by changing the maximum number of allowed iterations and/or the desired tolerance threshold.

\begin{table}
\centering
\caption{Examples of subgroups generated for various target sizes. The \textit{computed size} reports the expected subgroup size, and the \textit{actual size} represents the measured subgroup size for a sample of 10,000 instances.}
\label{tab:subgroup-examples}
\begin{tabular}{cccc}
\toprule
\textbf{Generated subgroup }& \textbf{   Target size   } & \textbf{   Computed size   } & \textbf{   Actual size   } \\ \midrule
\begin{tabular}[c]{@{}c@{}}\{ \textit{elevel} $\in [0, 3)\ \land$\\ \textit{zipcode} $\in [6, 7)\ \land$\\ \textit{age} $\in [29, 78)$ \}\end{tabular} & 0.05 & 0.0536 & 0.0552 \\ \midrule
\begin{tabular}[c]{@{}c@{}}\{ \textit{car} $\in [15, 19)\ \land$\\ \textit{salary} $\in [39000, 116000)\ \land$\\ \textit{zipcode} $\in [0, 8)$ \}\end{tabular} & 0.1 & 0.1045 & 0.107 \\\midrule
\begin{tabular}[c]{@{}c@{}}\{ \textit{zipcode} $\in [2, 5)\ \land$\\ \textit{salary} $\in [30000, 139000)\ \land$\\ \textit{age} $\in [22, 80)\ \land$\\ \textit{car} $\in [1, 20)$ \}\end{tabular} & 0.25 & 0.2505 & 0.2527 \\\midrule
\begin{tabular}[c]{@{}c@{}}\{ \textit{elevel} $\in [1, 4)\ \land$\\ \textit{age} $\in [20, 78)\ \land$\\ \textit{salary} $\in [21000, 140000)\ \land$\\ \textit{hyears} $\in [1, 30)$ \}\end{tabular} & 0.5 & 0.501 & 0.4965 \\ \bottomrule
\end{tabular}
\end{table}

\section{Experimental results}
\label{sec:results}
In this section, we show the performance of various drift detection techniques that are commonly adopted in literature on the proposed Subgroup Agrawal Drift dataset.
We are mainly interested in the change in performance of these techniques as the size of the drifting subgroup changes.

\paragraph{Drift detection techniques.} We considered the following drift detectors.
\begin{itemize}
\item \textbf{DDM (Drift Detection Method)}~\cite{gama2004learning} is a statistical technique that monitors the error rate of a model over time. 
When the error rate increases significantly, it indicates a possible change in the data distribution. 
If this increase surpasses a pre-defined drift threshold, DDM triggers the detection of a drift.
\item \textbf{EDDM (Early Drift Detection Method)}~\cite{baena2006early} improves upon DDM by focusing on the distance between errors instead of just the error rate.
This method aims to detect gradual changes more effectively. It calculates the average distance between errors and monitors the standard deviation of these distances. Significant changes in these metrics can indicate a drift, allowing the model to adapt more quickly to evolving data streams.
\item \textbf{HDDM (Hoeffding Drift Detection Method)}~\cite{frias2014online} is based on Hoeffding's inequality, which provides a way to determine the bounds of an estimator with high probability. 
HDDM uses this statistical method to detect changes in the distribution of incoming data compared to older data.
By comparing the distributions of recent data to older data, HDDM can identify when a significant change has occurred, suggesting that the underlying data distribution has drifted.
\item \textbf{FHDDM (Fast Hoeffding Drift Detection Method)}~\cite{pesaranghader2016fast} is an enhanced version of HDDM, designed to provide faster and more accurate detection. 
It applies Hoeffding's bounds to smaller windows of data, allowing it to detect drifts more quickly and with fewer false alarms. FHDDM is particularly useful in scenarios requiring rapid adaptation to changing data.


\end{itemize}

For each method, we identify the best-performing configuration of hyperparameters through a grid search on the dataset.

\paragraph{Dataset.} We adopt the proposed synthetic dataset for benchmarking drift detection techniques as the drifting subgroup sizes vary.
In particular, we are interested in the performance when the drifting subgroups are small, as these are the drifts that are intuitively more likely to go undetected.
We sample subgroup sizes from 1\% to 100\% (i.e., the full population) logarithmically. 

For each subgroup size, we conduct 100 experiments. For half of them, we inject drift to a random subgroup of the desired size (positive experiments).
The other half is instead not injected with any drift (negative experiments).

For positive experiments, we randomly choose one out of the 10 classification functions for the original concept, and a different one for the drift concept.
For negative experiments, we instead use a single concept throughout the entire experiment. 
For all experiments, we build a training set comprised of 10,000 points sampled from the underlying distribution and associated with the original concept.
We train a simple decision tree classification model with a depth of up to 5 nodes on this training set.
Subsequently, we sample 200 batches of data (1,000 points each). 
For positive experiments, the concept drift is injected gradually, as detailed in Subsection \ref{ssec:agrawal}.
The injection is centered around the $100^{th}$ batch, with a width of 100 batches. 
The subgroup accuracy in Figure \ref{fig:example} provides a visual intuition of the setting. 
We introduce a perturbation of 25\% of the input attributes to make the classification problem non-trivial.
For each experiment, the various drift detection techniques are used to monitor and potentially detect drifts.
Since each algorithm can potentially produce multiple drift detections, we count the number of detections. 
We determine the threshold on the minimum number of detections to trigger a drift alert using a ROC curve computed on 30\% of the experiments.
We use the rest of the experiments to compute the performance in terms of accuracy, $F_1$ score, False Positive Rate (FPR) and False Negative Rate (FNR), of various drift detection techniques. 

\begin{figure}[hbt!]
    \centering
    \includegraphics[width=\linewidth]{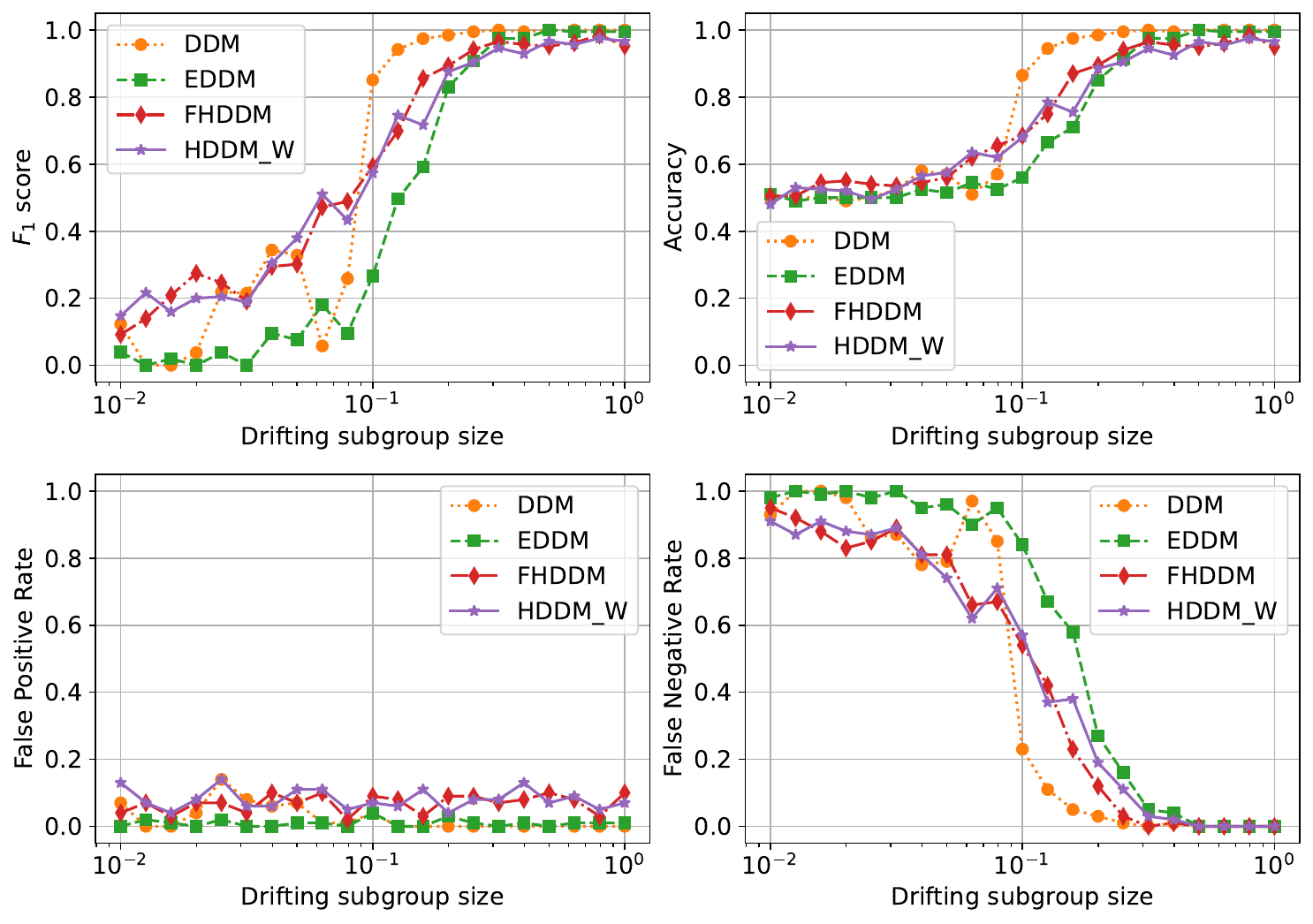}
    \caption{Performance, in terms of $F_1$ score, accuracy, False Positive Rate, and False Negative Rate, of various drift detection techniques on the binary task of detecting the occurrence of drift events. Subgroup Agrawal Drift dataset, perturbation of 25\%, various sizes of drifting subgroups. }
    \label{fig:results}
\end{figure}

\paragraph{Results.} 
Figure~\ref{fig:results} summarizes the main results. 
Both accuracy and $F_1$ highlight how all considered techniques achieve near-perfect performance in detecting drifts when the drifting subgroup is large enough (approximately 10\% of the dataset or more). 
Instead, none of the approaches achieved satisfactory results for lower support sizes. 
To better understand the cause of this drop in performance, we additionally computed the FPR and FNR for each technique for various sizes of drifting subgroups.

Interestingly, the FPR is largely unaffected by the size of the drifting subgroup. In other words, none of the considered approaches produces an excess of false positive predictions when smaller subgroups are drifting. 
This is in accordance with what was expected: drifts of smaller subgroups go unnoticed, meaning that fewer positive predictions are produced overall. 

Instead, the FNR plot presents a different perspective. 
In this case, it is clear that there exists an abundance of false negatives when the drifting subgroups are smaller in size. 
These false negatives are drifts that are not being detected: as expected, the various drift detection techniques cannot handle properly drifts of smaller subpopulations.

As the drifting subpopulations grow, the number of false positives produced decreases. Some approaches, such as DDM, have an earlier and sharper reduction in FNR, whereas other approaches (more significantly, EDDM) have a delayed response, meaning that they struggle to detect drifts even when the target subgroups are larger. 


\section{Conclusions}
\label{sec:conclusions} 
In this work, we highlighted a problem that affects commonly adopted drift detection techniques: drifts are only detected if they affect a large fraction of the original data. This implies that drifts affecting smaller subpopulations (e.g., minorities) may go undetected. This is problematic, since it implies that models may be silently drifting and underperforming for certain populations. 
To benchmark the performance of various detectors under subgroup drifts, we introduce the Subgroup Agrawal Drift Dataset, a synthetic data generator that injects a specific subgroup of a desired size with noise. 
The experimental results show indeed that commonly adopted techniques only detect subgroup drifts when these cover a large fraction of the dataset, producing a large number of false negatives in the case of smaller diverging subgroups. As a natural next step, we plan on addressing this shortcoming of current drift detection techniques. 

\section*{Acknowledgements}
This work is partially supported by the FAIR - Future Artificial Intelligence Research (PIANO NAZIONALE DI RIPRESA E RESILIENZA (PNRR) – MISSIONE 4 COMPONENTE 2, INVESTIMENTO 1.3 – D.D. 1555 11/10/2022, PE00000013) and the spoke ``FutureHPC \& BigData'' of the ICSC - Centro Nazionale di Ricerca in High-Performance Computing, Big Data and Quantum Computing, both funded by the European Union - NextGenerationEU. 
This manuscript reflects only the authors' views and opinions, neither the European Union nor the European Commission can be considered responsible for them.
\bibliographystyle{splncs04}
\bibliography{bibliography}

\begin{thebibliography}{1}
\providecommand{\url}[1]{\texttt{#1}}
\providecommand{\urlprefix}{URL }
\providecommand{\doi}[1]{https://doi.org/#1}

\bibitem{agrawal1993database}
Agrawal, R., Imielinski, T., Swami, A.: Database mining: A performance perspective. IEEE Transactions on Knowledge and Data Engineering  \textbf{5}(6),  914--925 (1993)

\bibitem{baena2006early}
Baena-Garc{\i}a, M., del Campo-{\'A}vila, J., Fidalgo, R., Bifet, A., Gavalda, R., Morales-Bueno, R.: Early drift detection method. In: {Fourth International Workshop on Knowledge Discovery from Data Streams}. vol.~6, pp. 77--86 (2006)

\bibitem{frias2014online}
Frias-Blanco, I., del Campo-{\'A}vila, J., Ramos-Jimenez, G., Morales-Bueno, R., Ortiz-Diaz, A., Caballero-Mota, Y.: Online and non-parametric drift detection methods based on hoeffding’s bounds. IEEE Transactions on Knowledge and Data Engineering  \textbf{27}(3),  810--823 (2014)

\bibitem{gama2004learning}
Gama, J., Medas, P., Castillo, G., Rodrigues, P.: Learning with drift detection. In: {Advances in Artificial Intelligence--SBIA 2004}. pp. 286--295 (2004)

\bibitem{montiel2021river}
Montiel, J., Halford, M., Mastelini, S.M., Bolmier, G., Sourty, R., Vaysse, R., Zouitine, A., Gomes, H.M., Read, J., Abdessalem, T., et~al.: River: machine learning for streaming data in python  (2021)

\bibitem{pesaranghader2016fast}
Pesaranghader, A., Viktor, H.L.: Fast hoeffding drift detection method for evolving data streams. In: ECML PKDD 2016. pp. 96--111 (2016)

\end{thebibliography}
\end{document}